\documentclass[sigconf]{aamas}

\usepackage{balance} % for balancing columns on the final page
\usepackage{multirow} 
\usepackage[table]{xcolor}
\usepackage{colortbl}
\usepackage{makecell}
\usepackage{graphicx}
\usepackage{subcaption} % 或者 subfig 宏包，建议 subcaption
\usepackage{enumitem}

\setcopyright{ifaamas}
\acmConference[AAMAS '26]{Proc.\@ of the 25th International Conference
on Autonomous Agents and Multiagent Systems (AAMAS 2026)}{May 25 -- 29, 2026}
{Paphos, Cyprus}{C.~Amato, L.~Dennis, V.~Mascardi, J.~Thangarajah (eds.)}
\copyrightyear{2026}
\acmYear{2026}
\acmDOI{}
\acmPrice{}
\acmISBN{}

\author{Shuai Wang}
\orcid{0009-0004-7323-0357}
\affiliation{
  \institution{Chalmers University of Technology}
  \city{Gothenburg}
  \country{Sweden}}
\email{shuaiwa@chalmers.se}

\author{Dhasarathy Parthasarathy}
\affiliation{
  \institution{Volvo Group}
  \city{Gothenburg}
  \country{Sweden}}
\email{dhasarathy.parthasarathy@volvo.com}

\author{Robert Feldt}
\affiliation{
  \institution{Chalmers University of Technology}
  \city{Gothenburg}
  \country{Sweden}}
\email{robert.feldt@chalmers.se}

\author{Yinan Yu}
\orcid{0000-0002-3221-7517}
\affiliation{
  \institution{Chalmers University of Technology}
  \city{Gothenburg}
  \country{Sweden}}
\email{yinan@chalmers.se}
\title{DomAgent: Leveraging Knowledge Graphs and Case-Based Reasoning for Domain-Specific Code Generation}

\begin{abstract}
Large language models (LLMs) have shown impressive capabilities in code generation. However, because most LLMs are trained on public domain corpora, directly applying them to real-world software development often yields low success rates, as these scenarios frequently require domain-specific knowledge. In particular, domain-specific tasks usually demand highly specialized solutions, which are often underrepresented or entirely absent in the training data of generic LLMs. To address this challenge, we propose DomAgent, an autonomous coding agent that bridges this gap by enabling LLMs to generate domain-adapted code through structured reasoning and targeted retrieval. A core component of DomAgent is DomRetriever, a novel retrieval module that emulates how humans learn domain-specific knowledge, by combining conceptual understanding with experiential examples. It dynamically integrates top-down knowledge-graph reasoning with bottom-up case-based reasoning, enabling iterative retrieval and synthesis of structured knowledge and representative cases to ensure contextual relevance and broad task coverage. DomRetriever can operate as part of DomAgent or independently with any LLM for flexible domain adaptation. We evaluate DomAgent on an open benchmark dataset in the data science domain (DS-1000) and further apply it to real-world truck software development tasks. Experimental results show that DomAgent significantly enhances domain-specific code generation, enabling small open-source models to close much of the performance gap with large proprietary LLMs in complex, real-world applications. The code is available at:
\url{https://github.com/Wangshuaiia/DomAgent}.
\end{abstract}
\keywords{Large Language Models, Domain-Specific Code Generation, Knowledge Graph, Agent-Based Systems}

\newcommand{\BibTeX}{\rm B\kern-.05em{\sc i\kern-.025em b}\kern-.08em\TeX}

\begin{document}

%%% The following commands remove the headers in your paper. For final 
%%% papers, these will be inserted during the pagination process.

\pagestyle{fancy}
\fancyhead{}

%%% The next command prints the information defined in the preamble.

\maketitle 

%%%%%%%%%%%%%%%%%%%%%%%%%%%%%%%%%%%%%%%%%%%%%%%%%%%%%%%%%%%%%%%%%%%%%%%%

\section{Introduction}
Recently, large language models (LLMs) have made remarkable progress in the field of code generation. Both closed-source models such as GPT-4o and Claude Sonnet 4.5, and open-source models like CodeLlama~\cite{roziere2023codellama}, Mistral~\cite{jiang2023mistral7b}, DeepSeek-Coder~\cite{guo2024deepseekcoder}, and Qwen-Coder~\cite{hui2024qwen25coder} have demonstrated impressive capabilities, and some of them even surpass human performance in general benchmark tasks. 
Tools like GitHub Copilot now leverage these models to provide adaptive, real-time coding assistance across various programming environments.

\begin{figure}[t]
    \centering
    \vspace{0.5cm}
    \includegraphics[width=1.0\linewidth]{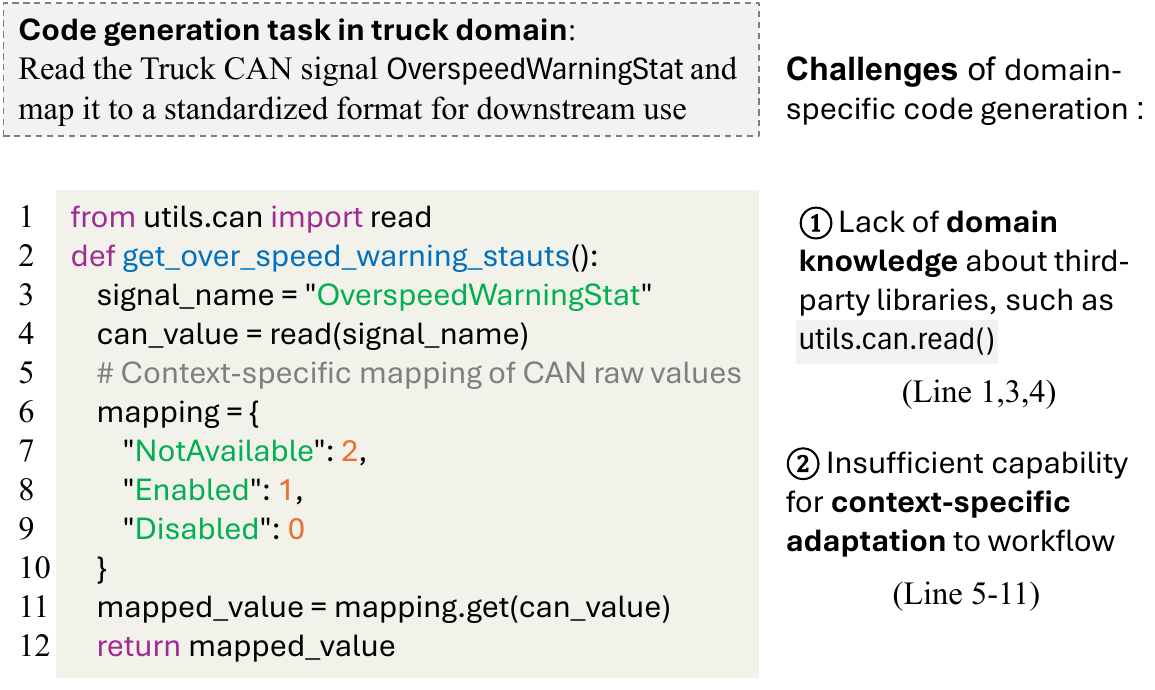}
    \caption{Example of a domain-specific code generation task in truck software development: reading the CAN signal \texttt{OverspeedWarningStat} and adapting the function with context-specific mapping to fit the workflow.}
    % Challenges of applying general LLMs to this domain include (1) lack of knowledge about third-party libraries and (2) insufficient capability for context-specific adaptation to the workflow.
    \label{fig:intro-example}
    \vspace{-0.5cm}
\end{figure}

In real-world applications, programming tasks are often highly domain-specific~\cite{li2024evocodebench,wang2025automating}. Software development across different industries, such as truck control systems or database management~\cite{hong2025next}, typically requires code that is tailored to the distinct requirements and constraints of each domain. While existing LLMs perform well on general-purpose, open-domain tasks, their performance degrades significantly when faced with specialized, domain-specific problems~\cite{gu2025effectiveness}.
One key reason is that LLMs often lack the required information to accomplish specialized tasks.
Domain-specific programming demands not only \textit{domain knowledge}, such as understanding how to use industry-specific libraries, but also the ability to produce \textit{specialized solutions} aligned with the operational logic of the target system. Figure~\ref{fig:intro-example} illustrates an example from truck software development: generating Python code to read the CAN signal \texttt{OverspeedWarningStat} and adapting the function with context-specific mapping to fit the workflow.
Directly using general LLMs poses two challenges: (1) a lack of knowledge about third-party libraries, and (2) insufficient capability for context-specific adaptation to the workflow.
% Successfully solving this task requires both identifying the correct CAN interface library and correctly transforming the raw signal into the required standard representation.

A straightforward way to address this challenge is to fine-tune LLMs with domain-specific data~\cite{anisuzzaman2025fine,ge2023supervised}. 
However, relying on fine-tuning to equip LLMs with domain-specific knowledge is often impractical, as fine-tuning introduces high development and maintenance costs~\cite{dong2024abilities}.  The resulting models often struggle to adapt to constantly updated libraries.
A more efficient and flexible solution is Retrieval-Augmented Generation (RAG)~\cite{lewis2020retrieval}, which augments existing LLMs with external knowledge without modifying model weights.
Popular software development frameworks such as AutoGen~\cite{wu2024autogen}, LangChain and TaskWeaver~\cite{qiao2023taskweaver} leverage this strategy to quickly integrate LLMs with domain knowledge at low cost.
Moreover, recent studies~\cite{ovadia2024fine,soudani2024fine} show that combining lightweight fine-tuning with RAG, using only a small number of domain examples, can substantially enhance model adaptability and accuracy for specialized programming tasks, achieving results comparable to much larger, fully fine-tuned systems.

In practice, RAG-based solutions dominate many real-world applications, accounting for over half (about 51\%) of deployed LLM-powered software systems~\cite{MenloVentures2024GenerativeAI}.
To provide \textit{domain knowledge} efficiently, recent work has shown that even basic retrieval methods like BM25 can significantly enhance performance by appending top-K results from external knowledge bases to prompts~\cite{yang2025empirical}. Using sentence embeddings for retrieval can also improve the retrieval accuracy~\cite{li2025retrieval,zeng2021pan}. However, excessively long retrieved content can negatively impact performance, as extended contexts may obscure key information and mislead the model~\cite{chang2024main}. Segmenting candidate texts into smaller chunks of 200–800 tokens is a way to solve this issue~\cite{wang-etal-2025-coderag}. \textit{Zhu et al.}~\cite{zhu2023acecoder} improved the segmenting by using structural cues such as abstract syntax trees (ASTs) to produce more meaningful splits~\cite{zhu2023acecoder}. Using another LLMs to filter out irrelevant information and compress the retrieved content also mitigate it~\cite{gao2024preference}. To introduce structured and concise knowledge, knowledge graphs (KGs) have also been explored as retrieval sources for code repair~\cite{ouyang2025knowledge}. KGs also contain the dependencies between software modules, supporting repository-level code generation~\cite{athale2025knowledge}. 
However, as these methods largely depend on plain textual similarity and neglect explicit links to relevant packages and functions, precisely retrieving relevant knowledge remains a fundamental challenge in complex real-world settings.

\textbf{Bottom-up retrieval:} For domain-specific tasks that require specialized knowledge and domain logic, a common approach is to leverage case-based reasoning (CBR), where a few-shot examples is included in the prompt to guide LLMs toward solving domain-adapted tasks~\cite{hatalis2025review,watson1994case}. 
Studies have shown that a fixed set of examples often fails to cover diverse scenarios. A more effective approach is to dynamically select examples based on the given task, using text embeddings to compute similarity scores~\cite{dannenhauer2024case}.
Subsequent work refined the retrieval process by re-ranking candidates through AST analysis to better capture code-level relationships~\cite{nashid2023retrieval}. 
To broaden the case coverage,  \textit{Tan et al.}~\cite{tan2024prompt} queried multiple information sources and integrated these retrieved contents through a segmented prompting strategy. Constructing high-quality case libraries from large training datasets is an effective way to improve case coverage~\cite{guo2024ds}. However, in real-world scenarios, creating a large and comprehensive case base is labor-intensive and time-consuming. Achieving high coverage with a small set of examples is still an important and practical challenge.

\textbf{Top-down retrieval:} Just as human learning also relies on structured understanding, rules, relationships, and conceptual hierarchies, LLMs require top-down knowledge to complement bottom-up example learning. KGs provide such structure by representing packages, functions, and their interrelations as nodes and edges. This structure makes it straightforward to link code examples to KG based on the package calling. For instance, as shown in Figure~\ref{fig:intro-example}, a code snippet calling \texttt{utils.can.read} can be explicitly connected to its corresponding node in the KG. Such links allow retrieving domain knowledge and locating relevant code examples beyond plain text similarity, by directly matching concrete package or function usage. In addition, explicitly evaluating the coverage of KG nodes by the packages and functions used in cases facilitates building a compact, diverse, and coverage-rich candidate case base.

Based on the above insights, we propose an agent system for domain-specific code generation that integrates both bottom-up (case-based) and top-down (knowledge-based) retrieval into a unified reasoning framework. The system is designed to emulate the human learning process: combining experiential reasoning from examples with conceptual understanding from structured knowledge. More specifically, the system leverages KGs as external repositories to provide accurate domain knowledge. To improve task coverage with a small case set, we exploit relational information in the KG to guide case selection, ensuring diversity by covering as many functions in the graph as possible. 
% To enable mutual enhancement between knowledge retrieval and case retrieval, we design a unified agent system powered by a reasoning LLM. Both processes are integrated into the LLM’s reasoning process, 
At the core of our system is a new retrieval module, DomRetriever, which dynamically integrates top-down and bottom-up retrieval to enhance both knowledge and case selection. It enables the model to iteratively identify, refine, and combine relevant knowledge with representative cases. Experiments on a benchmark dataset on data science domain show that our method significantly outperforms similar size LLMs. We also deploy the system in real-world truck software development tasks, demonstrating its robustness in practical settings.

Our contributions are as follows:

\begin{itemize}[topsep=1pt,itemsep=0.1pt]
\item We propose the first agent system (DomAgent) for domain-specific code generation that integrates structured knowledge (top-down) with case-based reasoning (bottom-up), improving both information retrieval and code generation.
\item We design a KG-guided case selction method that exploits structural relations in the knowledge graph to achieve broad task coverage with a small set of cases.
\item We develop a novel retrieval module (DomRetriever) that unifies knowledge-graph-guided retrieval and case-based reasoning into a dynamic, bidirectional process. DomRetriever can be seamlessly integrated into any LLM pipeline for flexible domain adaptation.
\item We validate the agent on a benchmark dataset and in real-world truck software development, demonstrating its effectiveness.
\end{itemize}

%%%%%%%%%%%%%%%%%%%%%%%%%%%%%%%%%%%%%%%%%%%%%%%%%%%%%%%%%%%%%%%%%%%%%%%%
\section{Related Work}
\subsection{Knowledge Retrieval for Code Generation}

RAG has become one of the most effective strategies to equip LLMs with external domain knowledge without modifying their parameters~\cite{lewis2020retrieval,wang-yu-2025-iquest,wang2025plugging}. By retrieving relevant information from external repositories, the model can access up-to-date and task-specific context to enhance code generation performance in specialized domains~\cite{zhao2024retrieval}. 
Yang et al.~\cite{yang2025empirical} conducted an empirical study showing that using basic retrieval methods, such as BM25, to fetch relevant content from external knowledge bases and simply appending the top-K results to the prompt can already yield strong performance. 
Li et al.~\cite{li2025retrieval} further enhanced retrieval accuracy by employing sentence embeddings for semantic search, enabling more precise matching between queries and relevant documents. However, long retrieved texts may overwhelm the model’s attention and obscure key information, leading to degraded generation quality~\cite{chang2024main}. To mitigate this issue, Wang et al.~\cite{wang-etal-2025-coderag} proposed splitting candidate texts into smaller chunks of 200–800 tokens, while Zhu et al.~\cite{zhu2023acecoder} leveraged structural cues from ASTs to produce more meaningful segmentations. Gao et al.~\cite{gao2024preference} introduced a secondary LLM to filter irrelevant content and compress retrieved texts, ensuring that only the most salient information is retained. Beyond plain-text retrieval, KGs offer a structured and concise way to represent domain knowledge. 
KGs can capture the dependencies between entities, which can be used to facilitate repository-level code generation. 
Ouyang et al.~\cite{ouyang2025knowledge} demonstrated that using KGs as retrieval sources can effectively provide relevant information for code repair tasks. 
Compared to unstructured text, KGs provide explicit relationships among entities such as packages, functions, and classes, making them particularly suitable for retrieving domain-specific programming knowledge.

\subsection{In-Context Learning and Case-Based Reasoning}
While retrieval-based methods focus on supplementing external knowledge, another effective strategy for domain-specific code generation is to provide task-specific guidance directly through examples. ICL~\cite{dong2024survey} and CBR~\cite{hatalis2025review,watson1994case} enable LLMs to learn from a few representative examples embedded in the prompt, helping them infer specialized problem-solving patterns without explicit model updates.
A common limitation of fixed few-shot prompts is their inability to generalize across diverse scenarios. Dannenhauer et al.~\cite{dannenhauer2024case} addressed this issue by building a case base and dynamically selecting examples based on semantic similarity, where both the query and cases were encoded into embeddings for retrieval. To further refine the selection, Nashid et al.~\cite{nashid2023retrieval} re-ranked candidate cases using AST analysis, which captures code-level structural relations that pure text similarity may miss. Tan et al.~\cite{tan2024prompt} proposed a segmented prompting strategy that integrates retrieved content from multiple sources to enhance contextual coverage. In addition, Guo et al.~\cite{guo2024ds} constructed a large and high-quality case base from extensive training data, and iteratively optimized it through selection and execution feedback to retain the most effective examples. Despite these advances, maintaining large case bases is labor-intensive and costly, making it impractical for fast-evolving industrial environments.

%%%%%%%%%%%%%%%%%%%%%%%%%%%%%%%%%%%%%%%%%%%%%%%%%%%%%%%%%%%%%%%%%%%%%%%%
\begin{figure*}[t]
    \centering
    \includegraphics[width=\textwidth]{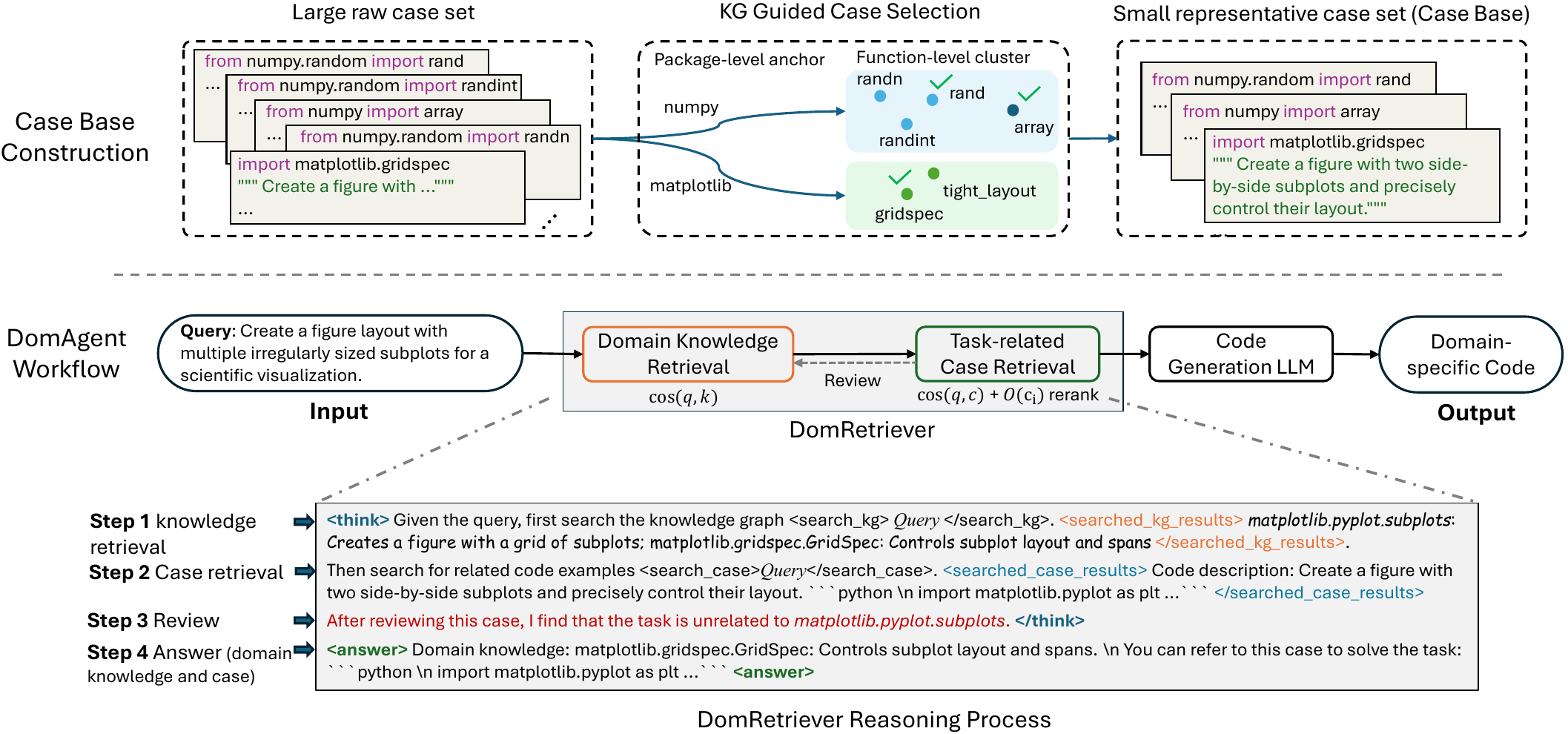} 
    \caption{Case base construction (top) and the overall workflow of our DomAgent (bottom). We first perform KG–guided selection to obtain representative cases. Then, our DomAgent retrieves domain knowledge from a KG and cases from the case base, while reviewing the retrieved domain knowledge in the reasoning process based on the selected case. Finally, using the answer of DomAgent, LLM generates the domain-specific code.}
    \label{fig:main}
\end{figure*}

\section{Problem defination}

We address the problem of domain-specific code generation by leveraging structured knowledge and reusable examples. 
Given a domain-specific generation task $q$, a KG $\mathcal{G}$, and a case base $\mathcal{B}$, 
the objective is to retrieve the necessary domain knowledge from $\mathcal{G}$ and identify relevant cases from $\mathcal{B}$, and then synthesize the target code $\hat{y}$ by integrating these retrieved resources.

Formally, we consider a KG $\mathcal{G} = \{\langle e, r, e' \rangle \mid e, e' \in \mathcal{E}, r \in \mathcal{R}\}$, where $\mathcal{E}$ and $\mathcal{R}$ represent the sets of entities and relations, respectively. Each triple $\langle e, r, e' \rangle$ encodes a relationship $r$ between entity $e$ and entity $e'$.  In our setting, entities represent domain knowledge such as packages, functions, their descriptions, and parameters, while edges represent direct relationships, e.g., a function belonging to a package.

The case base $\mathcal{B}$ is a collection of reference code cases. Its construction should ensure coverage and diversity while minimizing manual effort, since creating cases is costly and time-consuming. Our goal is to build a small but representative set of high-quality cases that achieves broad coverage.

\section{Methods}

Our approach, illustrated in Figure~\ref{fig:main}, comprises three key components. (1) \textit{Case Base Construction} module leverages the invocation relationships between functions and code cases in the knowledge graph to explicitly and efficiently sample representative cases; (2) \textit{Retrieval-Augmented Code Generation} module unifys knowledge and case retrieval through a reasoning LLM to enhance retrieval effectiveness; (3) For \textit{Agent Training}, we employ reinforcement learning to train the reasoning LLM to autonomously invoke retrieval tools, improving its ability to reason and access relevant knowledge when solving the coding task.

\subsection{Hierarchical Case Selection with Knowledge Graph Guidance}
\label{sec:sample}
To construct a representative case base that covers both diverse packages and their functional usage patterns, we leverage the hierarchical structure of the KG to perform hierarchical case selection.

\paragraph{Package-level anchoring.}
Suppose there are $n$ software packages $\mathcal{P}=\{p_1,p_2,\dots,p_n\}$ represented in the KG.  
For each package $p_i$, we take its \emph{root node} in the KG as an \textit{anchor} node $a_i$.
% Let $\mathcal{C}(a_i)=\{c_{i1},c_{i2},\dots\}$ denote the set of its child (sub-concept) nodes. 
Let $\mathcal{C}(a_i)=\{c_{i1},c_{i2},\dots\}$ denote the set of its child nodes (e.g., functions and attributes belonging to package $p_i$).
Each child node $c_{ij}$ is represented by its textual name $t_{ij}$ concatenated with its description $s_{ij}$, and then encoded into a continuous semantic vector space through an embedding function:
\begin{equation}
\mathbf{v}_{ij} = f_\mathrm{embed}(t_{ij} \,\Vert\, s_{ij}) \in \mathbb{R}^d
\label{eq:embedding}
\end{equation}
where $\Vert$ denotes string concatenation.

\paragraph{Function-level clustering.}
% 参考KBS
For each package $p_i$, we apply $k$-means clustering to its embedded child nodes:
\begin{equation}
\{\mathbf{v}_{ij}\} \xrightarrow[\text{k-means}]{}
\mathcal{K}_i = \{K_{i1},K_{i2},\dots,K_{im}\}
\end{equation}
where $m$ denotes the number of clusters per package.  
Each cluster $K_{ik}$ corresponds to a semantically coherent functional usage pattern.

\paragraph{Coverage-driven case selection.}
Each candidate code case $c$ imports some packages $p(x)$ and functions $F(x)$.  
Our goal is to build a case base $\mathcal{B}$ that:
1) covers as many packages as possible, and
2) maximizes the diversity of functional usage within each package.

Let
$
n' = |\{p(x)\mid x\in \mathcal{B}\}|, \,
m' = |\{K_{ik}\mid \exists x\in \mathcal{B},\,F(x)\cap K_{ik}\neq \emptyset\}|
$
be the number of distinct packages and the number of covered clusters in the current case base, respectively.
The \emph{coverage ratios} are then defined as
\begin{equation}
\alpha = \frac{n'}{n}, \quad \beta = \frac{m'}{m}
\end{equation}

We iteratively traverse the candidate code cases.  
For each candidate $x$, we test whether adding it to $\mathcal{B}$ increases either package coverage or cluster coverage:
\begin{equation}
\Delta \alpha = \frac{n'_\text{new}}{n}-\alpha, \quad
\Delta \beta = \frac{m'_\text{new}}{m}-\beta
\end{equation}
If $\Delta \alpha > 0$ or $\Delta \beta > 0$, the case is added to $\mathcal{B}$; otherwise, it is discarded as redundant.

\paragraph{Stopping criterion.}
We define two thresholds $\tau_1$ and $\tau_2$ to control the case selection process.  
The construction stops when
\begin{equation}
\alpha = \frac{n'}{n} > \tau_1
\quad\text{and}\quad
\beta = \frac{m'}{m} > \tau_2
\end{equation}
i.e., when both package-level and cluster-level coverage exceed predefined acceptable levels.  
This strategy ensures that each package is well represented while preserving the diversity of functional usage patterns within packages.

In addition, before storing the code to the case base, we execute it to ensure that it runs correctly.
We store these code cases in a vector database, where the key is the natural language description of the corresponding task to support similarity-based task retrieval.

\subsection{Domain-Specific Code Generation Agent}
\subsubsection{\textbf{DomRetriever}}
\label{sec:domretriever}

To generate domain-specific code, we retrieve both relevant domain knowledge and representative task-related cases.

\paragraph{Top-down retrieval.}
Given a task description $q$, we first determine which software package should be utilized. Specifically, we use an LLM as the classifier, and the inputs are both $q$ and the textual description $t_p$ of each candidate package. The LLM outputs a binary classification:
\begin{equation}
\hat{p}_i \;=\; LLM_P\!\big(q \,||\, t_p\big)\in\{0,1\}
\end{equation}
where $\hat{p}_i=1$ indicates that $p_i$ is applicable to the task $q$.

Then, we encode the task $q$  with $f_\mathrm{embed}$ to obtain $\mathbf{q}=f_\mathrm{embed}(q)\in\mathbb{R}^d$. Then we compute the cosine similarity between the encoded vector $f_{\mathrm{embed}}(q)$ and the embeddings of domain knowledge equations $\mathbf{v}_{ij}$ (as defined in Eq.~\ref{eq:embedding}):
\begin{equation}
s_{ij}(q)\;=\;\cos(\mathbf{q},\mathbf{v}_{ij}) \;=\; \frac{\mathbf{q}^\top \mathbf{v}_{ij}}{\|\mathbf{q}\|_2\,\|\mathbf{v}_{ij}\|_2}
\end{equation}
We select the nodes with the highest $T$ similarity scores as the relevant domain knowledge list $\mathcal{K} = \{k_1, k_2, \dots, k_{T}\}$.

\paragraph{Bottom-up retrieval.}
We first compute the semantic similarity between the task embedding $f_{\mathrm{embed}}(q)$ and the natural language descriptions of candidate code cases. The top-$R$ most similar cases are selected to form an initial case list $\mathcal{C} = \{c_1, c_2, \dots, c_{R}\}$. To refine this list, we further focus on the  similarity between cases and the retrieved packages (or functions, which we refer to as packages for simplicity in the following part). Specifically, for each case $c_i$, we compute the overlap count between the packages used in $c_i$ and those contained in $\mathcal{K}$:
\begin{equation}
O(c_i) = |\mathrm{Packages}(c_i) \cap \mathrm{Packages}(\mathcal{K})|
\end{equation}
Then we select the top case ${c}^*$ according to $O(c_i)$.

\paragraph{LLM-guided refinement with tool-use in the reasoning stage.}
To further enhance the relevance and completeness of the domain knowledge $\mathcal{K}$, we compare the selected case ${c}^*$ with $\mathcal{K}$, using a reasoning LLM as an agent to automatically refine $\mathcal{K}$ by removing irrelevant items or retrieving supplementary knowledge based on ${c}^*$. As illustrated in the example in Figure~\ref{fig:main}, during the review process, knowledge that is only superficially semantically related but functionally irrelevant can be filtered out.
To achieve this, both the domain knowledge retrieval and case retrieval processes are encapsulated as API tools, denoted as $\mathsf{SearchKG}(q)$ and $\mathsf{SearchCase}(q)$ respectively, which are invoked during the reasoning phase of the LLM.

As shown in Figure~\ref{fig:main}, we design two special tokens \texttt{<search\_kg>} and \texttt{<search\_case>} to trigger these searches respectively. 
% Retrieved results are wrapped using distinct symbols for clarity.
During the reasoning stage, similar to DeepSeek-R1~\cite{guo2025deepseek}, the LLM's internal reasoning is enclosed within the tag \texttt{<think>} $\cdots$ \texttt{</think>}.
In the subsequent answering stage (denoted by \texttt{<answer>} $\cdots$ \texttt{</answer>}),
the LLM outputs the finalized domain knowledge $\widehat{\mathcal{K}}$ together with the specialized solution $\hat{c}$.

\subsubsection{\textbf{Code Generation}}
We concatenate the original task description $q$, the refined domain knowledge $\widehat{\mathcal{K}}$, and the specialized solution $\hat{c}$ to form the final prompt, which is then fed into another $\mathrm{LLM}_{\mathrm{gen}}(\cdot)$ to generate the target code $\hat{y}$. Formally, the process can be expressed as:
\begin{equation}
\hat{y} = \mathrm{LLM}_{\mathrm{gen}}\big(\, q \, || \, \widehat{\mathcal{K}} \, || \, \hat{c} \big)
\end{equation}
In this module, we can either use the same reasoning LLM as DomRetriever for code generation or choose a more powerful LLM to assist with it.

% 这里需要给一个例子。

\subsection{Agent Training}

To enable the LLM to use retrieval tools, that is, to generate the special tokens \texttt{<search\_kg>} and \texttt{<search\_case>} for triggering retrieval, we constructed fine-tuning examples so that the LLM learns to produce these tokens. Specifically, we manually created several few-shot examples and concatenated them into long chain-of-thought (CoT) sequences. Then, given a query, we leveraged a powerful LLM (e.g., GPT-5) to generate reasoning steps and answers, like the reasoning process in Figure~\ref{fig:main}. Finally, we fine-tuned our LLM on this synthesized data to teach it to generate the retrieval-triggering tokens.

Based on this, we further apply reinforcement learning (RL) to guide the LLM to review the contents retrieved by the two tools and decide whether to filter out irrelevant knowledge. The final answer should include only the most relevant knowledge and cases. To provide a learning signal, we train a reward model \(f_{r}\) using a powerful external LLM. This model assesses the relevance of a query  \(q\) to the retrieved knowledge $\widehat{\mathcal{K}}$ and cases $\hat{c}$:
\begin{equation}
\mathrm{reward} = f_{r}\big(\, q \, || \, \widehat{\mathcal{K}} \, || \, \hat{c} \big).
\end{equation}

The obtained reward serves as the feedback signal for training the agent LLM. We employ the classical RL optimization algorithm GRPO~\cite{shao2024deepseekmath} to update the model parameters with respect to the reward function:
\begin{equation}
\theta^{*} = \arg\max_{\theta} \; \mathbb{E}_{q,\widehat{\mathcal{K}},\hat{c} \sim \pi_{\theta}} \big[ f_{r}(q \, || \, \widehat{\mathcal{K}} \, || \, \hat{c}) \big],
\end{equation}
where \(\pi_{\theta}\) denotes the policy of the LLM parameterized by \(\theta\).

\section{Experiments}
\subsection{Experimental Setup}
In this paper, we conduct experiments and ablation studies on two datasets: (1) the open benchmark dataset \emph{DS-1000}, and (2) a real-world domain-specific code generation dataset, \emph{Truck CAN Signal}.
\paragraph{Benchmark dataset.}
We first conducted experiments on the \textbf{DS-1000} benchmark dataset~\cite{lai2023ds} in the data science domain.
DS-1000 contains 1,000 tasks derived from 451 distinct Stack Overflow questions, covering seven widely used data science packages: \textit{NumPy, Pandas, Matplotlib, SciPy, scikit-learn, TensorFlow}, and \textit{PyTorch}.
The original questions are slightly modified to differ from their Stack Overflow sources, preventing LLMs from solving them by simply memorizing pre-training data. Because the tasks require calling and correctly using various functions across these domain libraries, they remain challenging even for state-of-the-art LLMs such as GPT-4o.

We use a knowledge graph \textbf{DS-KG} for data science packages which is constructed by Ouyang \textit{et al.}~\cite{ouyang2025knowledge}.
Each library function is represented as an entity, linking with attributes such as \textit{name} and \textit{description}.
They mined the official documentation of major data science libraries and obtained 505,640 triples in total.

\paragraph{Application in real-world truck software development.}
We further apply our agent for CAN signal reading and writing, which is a fundamental task in truck API development.
As shown in Figure~\ref{fig:intro-example}, working with CAN signals requires understanding third-party packages for CAN message access and applying specific signal transformations to ensure consistency across upstream and downstream systems. 
We leverage the truck manufacturer’s internal CAN signal documentation to provide domain knowledge.
In total, we use 776 CAN signals spanning six functional domains: \textit{Driver Productivity, Connected Systems, Energy, Vehicle Systems, Visibility}, and \textit{Dynamics}.

\paragraph{Implementation Details}
We conduct experiments using two LLMs as the backbone of our agent: \textit{LLaMA-3.1-8B-Instruct} and \textit{Qwen-2.5-7B}. We construct a set of 500 high-quality examples in open-domain task to teach the models how to invoke these retrieval tools. Once the model learns to use the retrieval tool, we apply it directly to our tasks. We sample only 300 examples (30\%) from the DS-1000 dataset to construct the case bases, and use the remaining 700 examples for testing.
We use sentence-BERT~\cite{reimers2019sentence} as the language encoding tool, with the hyperparameters $\tau_{1}$ and $\tau_{2}$ both set to 0.9.
The training is performed on 8 NVIDIA A100 80G GPUs in total. For each input query, we generate 16 outputs (rollouts). We train for 2 epochs with a batch size of 16 and a learning rate of $1\text{e}{-6}$. 
% The rollout temperature is set to 1, the PPO clip ratio is 0.2, and the KL divergence penalty coefficient is $1\text{e}{-5}$. 
DS-1000 provides predefined code contexts and testing functions for evaluation, while in the CAN signal reading and writing setting, engineers authored the corresponding test functions. Thus, we executed these test functions and adopted pass@1 as the metric.

% 我们的模型学会调用检索工具后，无需重复训练即可直接迁移领域

\subsection{Main Results}

\begin{table*}[htbp]
\centering
\caption{Performance comparison (pass@1) across different libraries and methods on Data Science dataset. All the results marked with an asterisk (*) are copied from the Magicoder paper~\cite{wei2024magicoder}.}
\resizebox{\textwidth}{!}{
\begin{tabular}{l c c c c c c c c c}
\hline
\textbf{Package} & \multirow{2}{*}{\textbf{Size}} 
& \textbf{Matplotlib} & \textbf{Numpy} & \textbf{Pytorch} & \textbf{Pandas} & \textbf{Scipy} & \textbf{Sklearn} & \textbf{Tensorflow} & \textbf{Total}\\
\textbf{Count} & & \textbf{155} & \textbf{220} & \textbf{291} & \textbf{68} & \textbf{106} & \textbf{115} & \textbf{45} & \textbf{1000}\\
\hline
\rowcolor{gray!30} \multicolumn{10}{l}{\textbf{Baseline (Vanilla LLMs)}} \\
Qwen2.5-7B                    & 7B & 54.2 & 34.9 & 25.3 & 16.8 & 23.1 & 21.4 & 29.7 & 29.3 \\
LLaMA3.1-8b                   & 8B & 55.1 & 36.2 & 26.4 & 18.3 & 24.2 & 22.5 & 30.1 & 30.4 \\
LLaMA3.3-70B                   & 70B  & 60.2 & 40.0 & 36.5 & 37.8 & 40.1 & 42.0 & 43.5 & 40.3 \\
GPT-3.5-Turbo & -  & 65.8 & 32.7 & 30.2 & 36.8 & 39.6 & 40.0 & 42.2 & 39.4 \\
GPT-4o        & -    & 65.2 & 56.8 & 41.9 & 47.1 & 48.1 & 50.4 & 46.7 & 51.0 \\
\hline
\rowcolor{gray!30} \multicolumn{10}{l}{\textbf{Coding Agents}} \\
StarCoder~\cite{listarcoder}     & 15B & 51.7$^{*}$  & 29.7$^{*}$  & 21.4$^{*}$  & 11.4$^{*}$  & 20.2$^{*}$  & \underline{29.5}$^{*}$  & 24.5$^{*}$  & 26.0$^{*}$ \\
WizardCoder~\cite{luo2024wizardcoder}   & 15B & 55.2$^{*}$  & 33.6$^{*}$  & 26.2$^{*}$  & 16.7$^{*}$  & 22.4$^{*}$  & 24.9$^{*}$  & 26.7$^{*}$  & 29.2$^{*}$ \\
CodeLLaMA-Python~\cite{roziere2023code} & 7B & 55.3$^{*}$  & 34.5$^{*}$  & 19.9$^{*}$  & 16.4$^{*}$  & 22.3$^{*}$  & 17.6$^{*}$  & 28.5$^{*}$  & 28.0$^{*}$ \\
WizardCoder-CL~\cite{wei2024magicoder}   & 7B & 53.5$^{*}$  & 34.4$^{*}$  & 25.7$^{*}$  & 15.2$^{*}$  & 21.0$^{*}$  & 24.5$^{*}$  & 28.9$^{*}$  & 28.4$^{*}$ \\
Magicoder-CL~\cite{wei2024magicoder}     & 7B & 54.6$^{*}$  & 34.8$^{*}$  & 24.7$^{*}$  & 19.0$^{*}$  & 25.0$^{*}$  & 22.6$^{*}$  & 28.9$^{*}$  & 29.9$^{*}$ \\
MagicoderS-CL (large-scale fine-tuning)~\cite{wei2024magicoder}     & 7B & 55.9$^{*}$  & \underline{40.6}$^{*}$  & \textbf{40.4}$^{*}$  & 28.4$^{*}$  & 28.8$^{*}$  & \textbf{35.8}$^{*}$  & \textbf{37.6}$^{*}$  & 37.5$^{*}$ \\
DomAgent (Qwen2.5-7B; \textbf{ours})  & 7B & \textbf{62.5} & 37.9 & 32.7 & \textbf{33.5} & \underline{30.6} & 29.1 & 36.9 & \underline{39.2} \\
DomAgent (LaMA3.1-8B; \textbf{ours})    & 8B & \textbf{62.5} & \textbf{43.1} & \underline{33.1} & \underline{29.2} & \textbf{31.5} & 29.3 & \underline{37.1} & \textbf{40.5} \\
\hline
\rowcolor{gray!30} \multicolumn{10}{l}{\textbf{External LLM (Code Gen) Enhanced with DomRetriever}} \\
LLaMA3.1-8B (DomRetriever) + LLaMA3.3-70B (Code Gen) & 70B  & 63.5 (+3.3) & 49.3 (+9.3) & 44.6 (+8.1) & 41.4 (+3.6) & 44.0 (+3.9) & 46.3 (+4.3) & 49.2 (+5.7) & 47.9 (+7.6) \\
LLaMA3.1-8B (DomRetriever) + GPT-4o (Code Gen)        & -    & 68.6 (+3.4) & 64.2 (+7.4) & 50.8 (+8.9) & 49.1 (+2.0) & 50.7 (+2.6) & 56.4 (+6.0) & 53.3 (+6.6) & 58.6 (+7.6) \\
\bottomrule
\end{tabular}
}
\label{tab:performance}
\end{table*}
\subsubsection{DS-1000}
Table \ref{tab:performance} presents a comprehensive comparison across multiple models and code-generation frameworks on the DS-1000 benchmark dataset. Among the vanilla LLMs, GPT-4o achieves the highest overall pass@1 accuracy (51.0\%), demonstrating strong general-purpose reasoning ability but revealing limited domain adaptation for specialized deep learning libraries, such as Pytorch and Tensorflow. Vanilla open-source models such as Qwen2.5-7B and LLaMA3.1-8B perform moderately, highlighting the performance disparity between proprietary and open models in zero-shot code generation settings. 

Within the coding-agent family, models like WizardCoder and Magicoder show clear improvements over standard LLMs through specialized fine-tuning on code-related data. However, even the strongest variant, MagicoderS-CL, achieves a total score of only 37.5\%, indicating that static code fine-tuning alone cannot fully generalize across diverse data science tasks. It is worth noting that MagicoderS-CL employs a two-stage large-scale fine-tuning strategy: it is initially trained on 75K carefully curated synthesized programming instruction data and subsequently fine-tuned on the 110K open-source complex instruction dataset Evol-Instruct. Such an approach requires large amounts of data and computational resources, making it impractical in real-world scenarios.

Our proposed DomAgent framework substantially improves the performance of small open-source models. DomAgent (Qwen2.5-7B) and DomAgent (LLaMA3.1-8B) reach 39.2\% and 40.5\%, respectively, outperforming all previous coding agents by approximately +2.0 to +3.0 percentage points. When paired with large external models for code generation, DomRetriever further enhances performance. For instance, coupling LLaMA3.1-8B (DomRetriever) with GPT-4o as the code generation model yields the best result at 58.6\%, representing a +7.6\% gain over GPT-4o alone. 

\subsubsection{Truck CAN Signal}
Table \ref{tab:volvo} compares the performance of large proprietary GPT-4o and small open-source models  Qwen2.5-7B across six highly specialized domains. Despite its strong general capability, the proprietary GPT-4o performs unsatisfactorily on these highly domain-specific code generation tasks, reaching an overall score of 71.22\%. When enhanced with DomRetriever, its performance increases to 98.04\%, highlighting the effectiveness of DomRetriever. 
This also ensures that our agent can operate effectively within real-world truck software development workflows.
In addition, the DomAgent, built on the small open-source Qwen2.5-7B model, achieves 96.64\%, nearly matching GPT-4o with DomRetriever and vastly outperforming its vanilla baseline (39.62\%). These results demonstrate that domain-adaptive architectures such as DomAgent can effectively bridge the performance gap between proprietary large models and small open-source models in complex, real-world industrial tasks.

\begin{table}[t]
\centering
\caption{Comparison of small \textbf{open models}-based DomAgent (Qwen2.5-7B-based) against GPT-4o-based models for domain-specific code generation tasks: truck CAN signal reading/writing across six domains.}
\renewcommand{\arraystretch}{1.15}
\resizebox{0.47\textwidth}{!}{
\begin{tabular}{m{2.5cm}cc|cc}
\hline
\multirow{2}{*}{\textbf{Domain}} &
\multicolumn{2}{c|}{\textbf{GPT-4o-based}} &
\multicolumn{2}{c}{\textbf{Qwen2.5-7B-based (open-source)}} \\
\cline{2-5}
& \textbf{Vanilla} & \makecell{\textbf{+DomRetriever}\\\textbf{(Qwen2.5-7B)}} 
& \textbf{Vanilla} & \textbf{DomAgent} \\
\hline
Driver productivity & 65.1 & 100 & 32.6 & 100 \\
Connected systems   & 73.4 & 97.4  & 46.8 & 97.8  \\
Energy              & 58.8 & 100 & 36.2 & 100 \\
Vehicle system      & 71.1 & 97.1  & 35.1 & 94.3  \\
Visibility          & 82.2 & 100 & 44.3 & 100 \\
Dynamics            & 75.5 & 96.4  & 39.8 & 91.4  \\
\hline
\textbf{Total} & \textbf{71.22} & \textbf{98.04} & \textbf{39.62} & \textbf{96.64} \\
\hline
\end{tabular}}
\label{tab:volvo}
\end{table}

\begin{table*}[htbp]
\centering
\caption{Ablation study of different modules on the Data Science dataset.}
\resizebox{\textwidth}{!}{
\begin{tabular}{l c c c c c c c c c}
\hline
\multirow{2}{*}{\textbf{Count}} & \multirow{2}{*}{\textbf{Size}} 
& \textbf{Matplotlib} & \textbf{Numpy} & \textbf{Pytorch} & \textbf{Pandas} & \textbf{Scipy} & \textbf{Sklearn} & \textbf{Tensorflow} & \textbf{Total}\\
& & \textbf{155} & \textbf{220} & \textbf{291} & \textbf{68} & \textbf{106} & \textbf{115} & \textbf{45} & \textbf{1000}\\
\hline
\rowcolor{gray!30} \multicolumn{10}{l}{\textbf{Ablation on End-to-end DomAgent}} \\
Qwen2.5-7B                    & 7B & 54.2 & 34.9 & 25.3 & 16.8 & 23.1 & 21.4 & 29.7 & 29.3 \\
Qwen2.5-7B+KG                 & 7B & 56.1 (+1.9) & 37.5 (+2.6) & 27.8 (+2.5) & 18.4 (+1.6) & 25.2 (+2.1) & 23.5 (+2.1) & 31.2 (+1.5) & 31.4 (+2.1) \\
Qwen2.5-7B+CBR                & 7B & 57.3 (+3.1) & 32.9 (-2.0) & 32.4 (+7.1) & 24.6 (+7.8) & 27.5 (+4.4) & 28.5 (+7.1) & 37.9 (+8.2) & 34.8 (+5.5) \\
Qwen2.5-7B+KG+CBR (RAG pipeline)  & 7B & 62.3 (+8.1) & 40.9 (+6.0) & 31.6 (+6.3) & 23.8 (+7.0) & 30.4 (+7.3) & 27.3 (+5.9) & 36.3 (+6.6) & 37.4 (+8.1) \\
Qwen2.5-7B+KG+CBR (DomAgent)  & 7B & 62.5 (+8.3) & 37.9 (+3.0) & 32.7 (+7.4) & 33.5 (+16.7) & 30.6 (+7.5) & 29.1 (+7.7) & 36.9 (+7.2) & 39.2 (+9.9) \\
\hline
LLaMA3.1-8b                   & 8B & 55.1 & 36.2 & 26.4 & 18.3 & 24.2 & 22.5 & 30.1 & 30.4 \\
LLaMA3.1-8b+KG                & 8B & 57.2 (+2.1) & 38.1 (+1.9) & 28.6 (+2.2) & 19.1 (+0.8) & 26.0 (+1.8) & 24.0 (+1.5) & 32.5 (+2.4) & 32.2 (+1.8) \\
LLaMA3.1-8b+CBR               & 8B & 61.6 (+6.5) & 40.1 (+3.9) & 30.8 (+4.4) & 26.6 (+8.3) & 30.4 (+6.2) & 28.7 (+6.2) & 37.4 (+7.3) & 37.6 (+7.2) \\
LLaMA3.1-8b+KG+CBR (RAG pipeline) & 8B & 60.4 (+5.3) & 41.3 (+5.1) & 32.6 (+6.2) & 24.1 (+5.8) & 30.6 (+6.4) & 29.1 (+6.6) & 36.7 (+6.6) & 39.2 (+8.8) \\
LLaMA3.1-8b+KG+CBR (DomAgent)    & 8B & 62.5 (+7.4) & 43.1 (+6.9) & 33.1 (+6.7) & 29.2 (+10.9) & 31.5 (+7.3) & 29.3 (+6.8) & 37.1 (+7.0) & 40.5 (+10.1) \\
\hline
\rowcolor{gray!30} \multicolumn{10}{l}{\textbf{Ablation on External LLM (Code Gen) Enhanced with DomRetriever}} \\
LLaMA3.3-70B                   & 70B  & 60.2 & 40.0 & 36.5 & 37.8 & 40.1 & 42.0 & 43.5 & 40.3 \\
LLaMA3.3-70B+KG                & 70B  & 61.0 (+0.8) & 42.2 (+2.2) & 37.0 (+0.5) & 38.4 (+0.6) & 41.0 (+0.9) & 42.7 (+0.7) & 44.2 (+0.7) & 40.4 (+0.1) \\
LLaMA3.3-70B+CBR               & 70B  & 62.5 (+2.3) & 47.1 (+7.1) & 38.5 (+2.0) & 40.0 (+2.2) & 42.5 (+2.4) & 44.1 (+2.1) & 45.5 (+2.0) & 46.0 (+5.7) \\
LLaMA3.3-70B+KG+CBR (RAG pipeline) & 70B  & 63.1 (+2.9) & 47.8 (+7.8) & 39.1 (+2.6) & 40.6 (+2.8) & 43.2 (+3.1) & 44.9 (+2.9) & 46.2 (+2.7) & 46.5 (+6.2) \\
Qwen2.5-7B (DomRetriever) + LLaMA3.3-70B (Code Gen)         & 70B  & 63.9 (+3.7) & 48.2 (+8.2) & 43.8 (+7.3) & 41.8 (+4.0) & 44.3 (+4.2) & 46.0 (+4.0) & 47.3 (+3.8) & 47.4 (+7.1) \\
LLaMA3.1-8B (DomRetriever)  + LLaMA3.3-70B (Code Gen)         & 70B  & 63.5 (+3.3) & 49.3 (+9.3) & 44.6 (+8.1) & 41.4 (+3.6) & 44.0 (+3.9) & 46.3 (+4.3) & 49.2 (+5.7) & 47.9 (+7.6) \\
\hline
GPT-4o        & -    & 65.2 & 56.8 & 41.9 & 47.1 & 48.1 & 50.4 & 46.7 & 51.0 \\
GPT-4o+KG        & -    & 65.5 (+0.3) & 56.9 (+0.1) & 41.3 (-0.6) & 46.2 (-0.9) & 50.1 (+2.0) & 51.8 (+1.4) & 52.0 (+5.3) & 51.2 (+0.2) \\
GPT-4o+CBR        & -    & 69.0 (+3.8) & 60.8 (+4.0) & 49.1 (+7.2) & 52.0 (+4.9) & 53.2 (+5.1) & 53.7 (+3.3) & 53.5 (+6.8) & 56.1 (+5.1) \\
GPT-4o+KG+CBR (RAG pipeline)      & -    & 68.5 (+3.3) & 61.2 (+4.4) & 50.0 (+8.1) & 50.6 (+3.5) & 51.0 (+2.9) & 56.4 (+6.0) & 53.1 (+6.4) & 56.4 (+5.4) \\
Qwen2.5-7B (DomRetriever) + GPT-4o (Code Gen) & -    & 67.8 (+2.6) & 62.3 (+5.5) & 49.3 (+7.4) & 49.6 (+2.5) & 50.3 (+2.2) & 55.6 (+5.2) & 52.2 (+5.5) & 57.8 (+6.8) \\
LLaMA3.1-8B (DomRetriever) + GPT-4o (Code Gen) & -    & 68.6 (+3.4) & 64.2 (+7.4) & 50.8 (+8.9) & 49.1 (+2.0) & 50.7 (+2.6) & 56.4 (+6.0) & 53.3 (+6.6) & 58.6 (+7.6) \\
\bottomrule
\end{tabular}
}
\label{tab:performance}
\end{table*}

\begin{table*}[htbp]
\centering
\caption{Ablation study on the truck CAN signal code generation task.}
\resizebox{\textwidth}{!}{
\renewcommand{\arraystretch}{1.2}
\begin{tabular}{lccccccc}
\hline
\textbf{Domain} &
\makecell{\textbf{Driver}\\\textbf{productivity}} &
\makecell{\textbf{Connected}\\\textbf{systems}} &
\textbf{Energy} &
\makecell{\textbf{Vehicle}\\\textbf{system}} &
\textbf{Visibility} &
\textbf{Dynamics} &
\textbf{Total} \\
\textbf{Count} & \textbf{135} & \textbf{212} & \textbf{72} & \textbf{136} & \textbf{56} & \textbf{155} & \textbf{776} \\
\hline
\rowcolor{gray!30} \multicolumn{8}{l}{\textbf{Ablation on End-to-end DomAgent}} \\
Qwen2.5-7B & 32.6 & 46.8 & 36.2 & 35.1 & 44.3 & 39.8 & {39.62} \\
Qwen2.5-7B + KG & 62.1 (+29.5) & 76.4 (+29.6) & 78.7 (+42.5) & 67.3 (+32.2) & 65.4 (+21.1) & 72.5 (+32.7) & {70.89 (+31.27)} \\
Qwen2.5-7B + CBR & 74.4 (+41.8) & 88.1 (+41.3) & 86.3 (+50.1) & 84.8 (+49.7) & 89.3 (+45.0) & 85.9 (+46.1) & {84.57 (+44.95)} \\
Qwen2.5-7B+KG+CBR (RAG pipeline) & 92.2 (+59.6) & 91.1 (+44.3) & 94.2 (+58.0) & 87.1 (+52.0) & 95.2 (+50.9) & 90.4 (+50.6) & {91.03 (+51.41)} \\
Qwen2.5-7B+KG+CBR (DomAgent) & 100 (+67.4) & 97.8 (+51.0) & 100 (+63.8) & 94.3 (+59.2) & 100 (+55.7) & 91.4 (+51.6) & {96.64 (+57.02)} \\

\hline
\rowcolor{gray!30} \multicolumn{8}{l}{\textbf{Ablation on External LLM (Code Gen) Enhanced with DomRetriever}} \\
GPT-4o & 65.1 & 73.4 & 58.8 & 71.1 & 82.2 & 75.5 & {71.22} \\
GPT-4o + KG & 85.2 (+20.1) & 85.3 (+11.9) & 84.3 (+25.5) & 90.7 (+19.6) & 94.9 (+12.7) & 90.0 (+14.5) & {87.80 (+16.58)} \\
GPT-4o + CBR & 93.5 (+28.4) & 92.1 (+18.7) & 90.2 (+31.4) & 88.3 (+17.2) & 93.4 (+11.2) & 89.7 (+14.2) & {91.10 (+19.88)} \\
GPT-4o+KG+CBR (RAG pipeline) & 97.6 (+32.5) & 96.2 (+22.8) & 96.5 (+37.7) & 95.1 (+24.0) & 98.0 (+15.8) & 96.6 (+21.1) & {96.49 (+25.27)} \\
Qwen2.5-7B (DomRetriever) + GPT-4o (Code Gen) & 100 (+34.9) & 97.4 (+24.0) & 100 (+41.2) & 97.1 (+26.0) & 100 (+17.8) & 96.4 (+20.9) & {98.04 (+26.82)} \\
LLaMA3.1-8B (DomRetriever) + GPT-4o (Code Gen) & 100 (+34.9) & 98.3 (+24.9) & 100  (+41.2) & 98.6 (+27.5) & 100 (+17.8) & 97.2 (+21.7) & 98.71 (+27.49) \\
\bottomrule
\end{tabular}
}
\label{tab:ablation_performance_truck}
\end{table*}

\subsection{Ablation Study}
We formulate our ablation design to understand how each component in our proposed framework contributes to domain-specific code generation across diverse data science libraries and real-world expert domains.

\subsubsection{DS-1000}
We first compare the configurations with knowledge grounding (\textit{+KG}) and case-based reasoning (\textit{+CBR}), and their combination with a standard RAG mechanism \textit{+KG+CBR (RAG pipeline)} to quantify how each component contributes individually and jointly. The results show that while knowledge grounding improves baseline performance, case-based reasoning provides larger gains. The combination of \textit{+KG+CBR} in a standard RAG pipeline further enhances overall performance, with improvements roughly additive to the individual contributions. 
Further, comparing \textit{+KG+CBR (RAG pipeline)} and \textit{+KG+CBR (DomAgent)} reveals that DomAgent’s retrieval mechanism (cf. Section~\ref{sec:domretriever}) consistently yield improvements across both Qwen2.5-7B and LLaMA3.1-8B, achieving competitive results with \textit{LLaMA3.3-70B}, particularly in specialized tasks requiring deeper logical composition and multi-step data manipulation, such as operations in Pandas and Tensorflow. Moreover, we study the synergy between \textit{DomRetriever} and large external code-generation models. Combining small retrieval models (e.g., LLaMA3.1-8B or Qwen2.5-7B) with strong generators (e.g., GPT-4o or LLaMA3.3-70B) consistently improves performance across libraries.

\begin{figure}[t]
    \centering
    \includegraphics[width=1.0\linewidth]{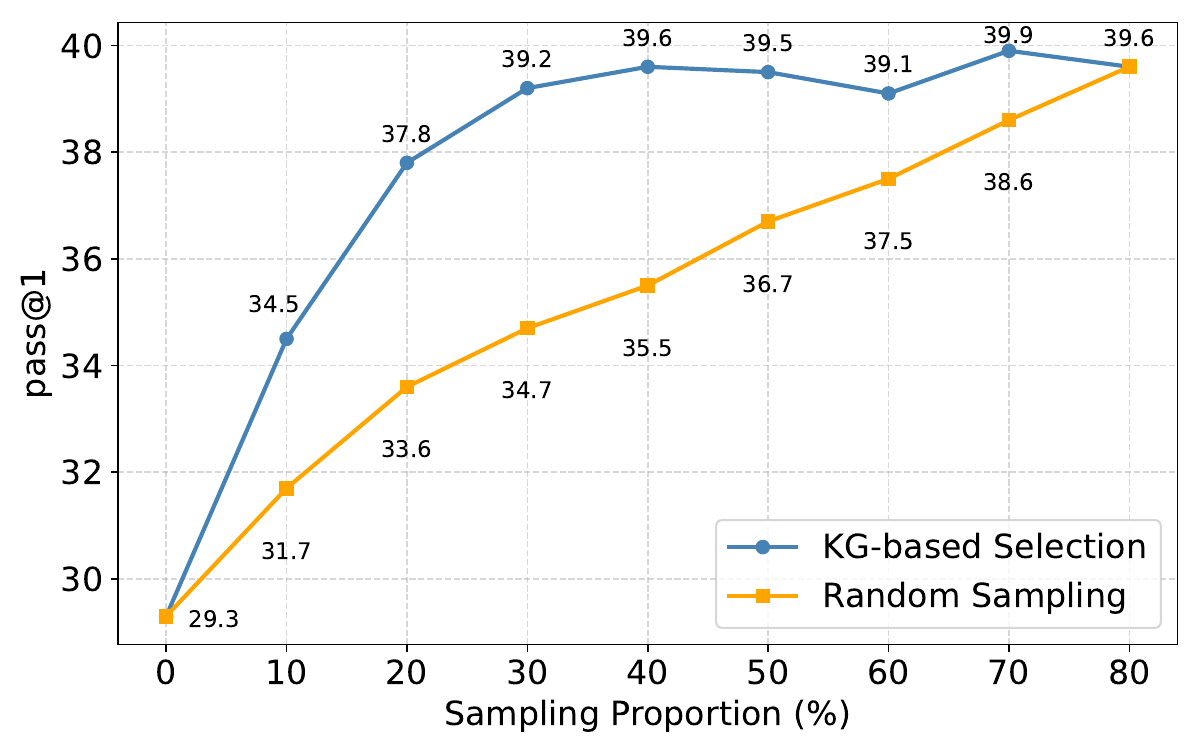}
    \caption{Effect of KG-based case selection versus random sampling with different sampling proportions on case base construction, using the Qwen2.5-7B-based DomAgent on the \textit{Data Science} dataset.}
    \label{fig:sampling-comparison}
\end{figure}

Case-based reasoning is an important step in DomAgent.
Since constructing a high-quality case base is both expensive and time-consuming, an efficient strategy is crucial.
To evaluate the effectiveness of our Hierarchical Case Selection with Knowledge Graph Guidance introduced in Section~\ref{sec:sample}, we conduct an ablation study by gradually increasing the proportion of sampled cases. As shown in Figure~\ref{fig:sampling-comparison}, performance (pass@1) steadily improves with larger sampling proportions for both strategies, but the KG-guided case selection consistently outperforms random sampling across all proportions. The improvement is most pronounced in the low-sampling regime (achieving near-full performance with only 30\% of the data), demonstrating that knowledge-guided selection effectively identifies more representative and contextually diverse examples. 

\subsubsection{Truck CAN Signal}

We conducted experiments on six types of truck CAN signals using our proposed DomAgent, which employs Qwen2.5-7B as its backbone. We also tested an alternative configuration that leverages GPT-4o as an external LLM for code generation. The results are summarized in Table~\ref{tab:ablation_performance_truck}.

The results clearly demonstrate that DomAgent significantly outperforms the vanilla models in both settings. Specifically, DomAgent with Qwen2.5-7B achieves a 57.02\% improvement over the vanilla Qwen2.5-7B model, while DomAgent with GPT-4o achieves a 26.82\% improvement. Overall, both models achieve strong final performance: Qwen2.5-7B reaches a pass@1 of 96.64, and GPT-4o reaches 98.04. Notably, DomAgent performs consistently well across all six CAN signal types. Both configurations even achieve 100\% pass@1 on the \emph{Driver Productivity}, \emph{Energy}, and \emph{Visibility} signal types, confirming the reliability of our method for industrial deployment.

By contrast, the vanilla models perform poorly, even though we manually curated representative cases and carefully tuned prompts. The vanilla Qwen2.5-7B only achieves 39.62 pass@1, while the stronger GPT-4o reaches 71.22 pass@1. This finding highlights the limitations of relying solely on fixed prompt engineering or static few-shot examples. Without injecting external domain knowledge and customizing the reasoning process, existing LLMs struggle to handle domain-specific code generation tasks effectively.

When we add domain knowledge (\textit{+KG}), Qwen2.5-7B improves by 31.27 points, and GPT-4o improves by 16.58 points, both showing substantial gains. This improvement is mainly because our knowledge graph is constructed from internal CAN signal documentation, which contains descriptions of signal attributes and natural-language explanations of how to use essential third-party libraries. With this domain-specific knowledge, the models can accurately call internal CAN signal read/write tools. Interestingly, GPT-4o achieves 87.80 pass@1 using only \textit{+KG}, reflecting its strong document understanding and code generation capabilities, allowing it to leverage documentation effectively even without explicit few-shot examples.

Adding case-based reasoning (\textit{+CBR}) leads to an even larger improvement: Qwen2.5-7B +KG improves by an additional 44.95 points, and GPT-4o improves by 19.88 points. This is because retrieving highly relevant cases enables more effective few-shot learning, where the retrieved cases often include domain knowledge, such as concrete examples of how to invoke third-party libraries or convert CAN signals.

Comparing the \textit{RAG pipeline} with our DomAgent shows that review and filtering of irrelevant information further enhance performance. Moreover, reinforcement learning within DomAgent strengthens its ability to autonomously explore and reason, contributing to its superior results.

Overall, each module of our approach brings substantial improvements to domain-specific code generation. The integration of domain knowledge, case-based reasoning, information filtering, and reinforcement learning enables DomAgent to achieve state-of-the-art results and makes it feasible for deployment in real industrial environments.

\section{Conclusion and Future Work}
In this work, we introduced DomAgent, an autonomous agent for domain-specific code generation that leverages KG–based retrieval to accurately acquire structured domain knowledge. During case retrieval, DomAgent exploits the relationships between packages and cases, re-ranking candidates based on package overlap. By integrating the reasoning capabilities of LLMs, DomAgent can invoke retrieval tools during the reasoning process, review the retrieved domain knowledge against the target case, and filter out irrelevant information to improve retrieval precision. We trained the LLM to use retrieval tools through supervised fine-tuning and employed reinforcement learning to encourage the model to explore the relevance between cases and domain knowledge. In addition, we proposed a KG-guided case selection method for constructing the case base, achieving comparable performance by selecting only 30\% of the cases compared to random sampling with 80\%. DomAgent can serve as a retriever for essential domain knowledge and cases, making it easily integrable with other external LLMs for code generation. Experiments on both a benchmark dataset and a real-world truck domain dataset demonstrated that our approach outperforms LLMs of similar size. Moreover, we successfully deployed DomAgent in a real factory to generate code for accessing truck CAN signals. These results collectively validate the effectiveness and practical value of our method.

In future work, we plan to enhance DomAgent along three main directions. First, we will incorporate AST–based fine-grained similarity analysis \cite{zhang2025cast}, to improve the performance of retrieving relevant code cases at the structural level. Second, we will explore the use of soft prompts \cite{li-etal-2025-prompt}, a lightweight form of knowledge injection, to enable the model to efficiently adapt to new domains with minimal retraining. Finally, we intend to evaluate DomAgent in a wider range of real-world industrial scenarios to further validate its practicality and generalizability.

\section*{Acknowledgment}
This work was partially funded by the Autonomous Systems and Software Program (WASP), supported by the Knut and Alice Wallenberg Foundation, and the Chalmers Artificial Intelligence Research Centre (CHAIR).

%%%%%%%%%%%%%%%%%%%%%%%%%%%%%%%%%%%%%%%%%%%%%%%%%%%%%%%%%%%%%%%%%%%%%%%%

%%% The next two lines define, first, the bibliography style to be 
%%% applied, and, second, the bibliography file to be used.

\bibliographystyle{ACM-Reference-Format} 
\bibliography{sample}

%%%%%%%%%%%%%%%%%%%%%%%%%%%%%%%%%%%%%%%%%%%%%%%%%%%%%%%%%%%%%%%%%%%%%%%%

\end{document}